\documentclass[10pt,twocolumn,letterpaper]{article}

\usepackage{iccv}
\usepackage{times}
\usepackage{epsfig}
\usepackage{graphicx}
\usepackage{amsmath}
\usepackage{amssymb}

\usepackage{caption}
\usepackage{subcaption}
\usepackage{graphicx}
\usepackage{url}
\usepackage{amsmath}
\usepackage{wrapfig}
\usepackage{multirow}
\usepackage{mathrsfs}
\usepackage{booktabs}
\usepackage{bbm}

\usepackage[breaklinks=true,bookmarks=false]{hyperref}

\iccvfinalcopy 

\ificcvfinal\fi

\begin{document}

\title{Learning an Efficient Network for Large-Scale Hierarchical Object Detection with Data Imbalance: 3rd Place Solution to Open Images Challenge 2019}

\author{
Xingyuan Bu\\
Beijing Institute of Technology\\
{\tt\small buxingyuan@bit.edu.cn}\\
\and
Junran Peng\\
University of Chinese Academy of Sciences\\
{\tt\small pengjunran2015@ia.ac.cn}
\and
Changbao Wang\\
Beihang University\\
{\tt\small wangchangbao@buaa.edu.cn}
\and
Cunjun Yu\\
Nanyang Technological University\\
{\tt\small cyu002@e.ntu.edu.sg}
\and
Guoliang Cao\\
Peking University\\
{\tt\small caogl@pku.edu.cn}
}
\maketitle
\ificcvfinal\thispagestyle{empty}\fi

\begin{abstract}
This report details our solution to the Google AI Open Images Challenge 2019 Object Detection Track.
Based on our detailed analysis on the Open Images dataset,
it is found that there are four typical features: large-scale, hierarchical tag system, severe annotation incompleteness and data imbalance.
Considering these characteristics, many strategies are employed, including larger backbone, distributed softmax loss, class-aware sampling, expert model, and heavier classifier.
In virtue of these effective strategies, our best single model could achieve a mAP of 61.90.
After ensemble, the final mAP is boosted to 67.17 in the public leaderboard and 64.21 in the private leaderboard, which earns 3rd place in the Open Images Challenge 2019.
\end{abstract}

\section{Introduction}
Object detection has been a challenging issue in the field of computer vision for a long time.
It is a basic and important task for a variety of industrial applications, such as autonomous driving and sense analysis.
Due to the blooming development of deep learning and large-scale dataset, great progresses have been achieved in object detection in recent years~\cite{girshick2014rich, he2014spatial, girshick2015fast, ren2015faster, lin2017feature, he2017mask, dai2017deformable, liu2018path, cai2017cascade, lin2018focal, redmon2018yolov3}.
Open Images Dataset V5~\cite{openimages, OpenImagesSegmentation} is currently the largest object detection dataset.
Different from its predecessors, such as Pascal VOC~\cite{everingham2010pascal}, MS COCO~\cite{lin2014microsoft}, and Objects365~\cite{obj365}, Open Images dataset consists of extremely large-scale annotations including 12M bounding boxes for 500 categories on 1.7M images, as shown in Table~\ref{tab:dataset}.
\begin{table*}
\small
\centering
\begin{tabular}{lllll}
\toprule
Dataset     & Pascal VOC & COCO  & Objects365    & Open Images    \\
\midrule
Categories  & 20          & 80        & 365       & 500           \\
Images      & 11,540      & 123,287   & 638,630   & 1,784,662     \\
Boxes       & 27,450      & 886,287   & 10,101,056& 12,421,955    \\
Boxes/image & 2.4         & 7.2       & 15.8      & 7.0           \\
\bottomrule
\end{tabular}
\caption{The comparison of different detection datasets.}
\label{tab:dataset}
\end{table*}

\begin{figure}[t]
\begin{minipage}{\linewidth}
\begin{center}
\includegraphics[width=\linewidth]{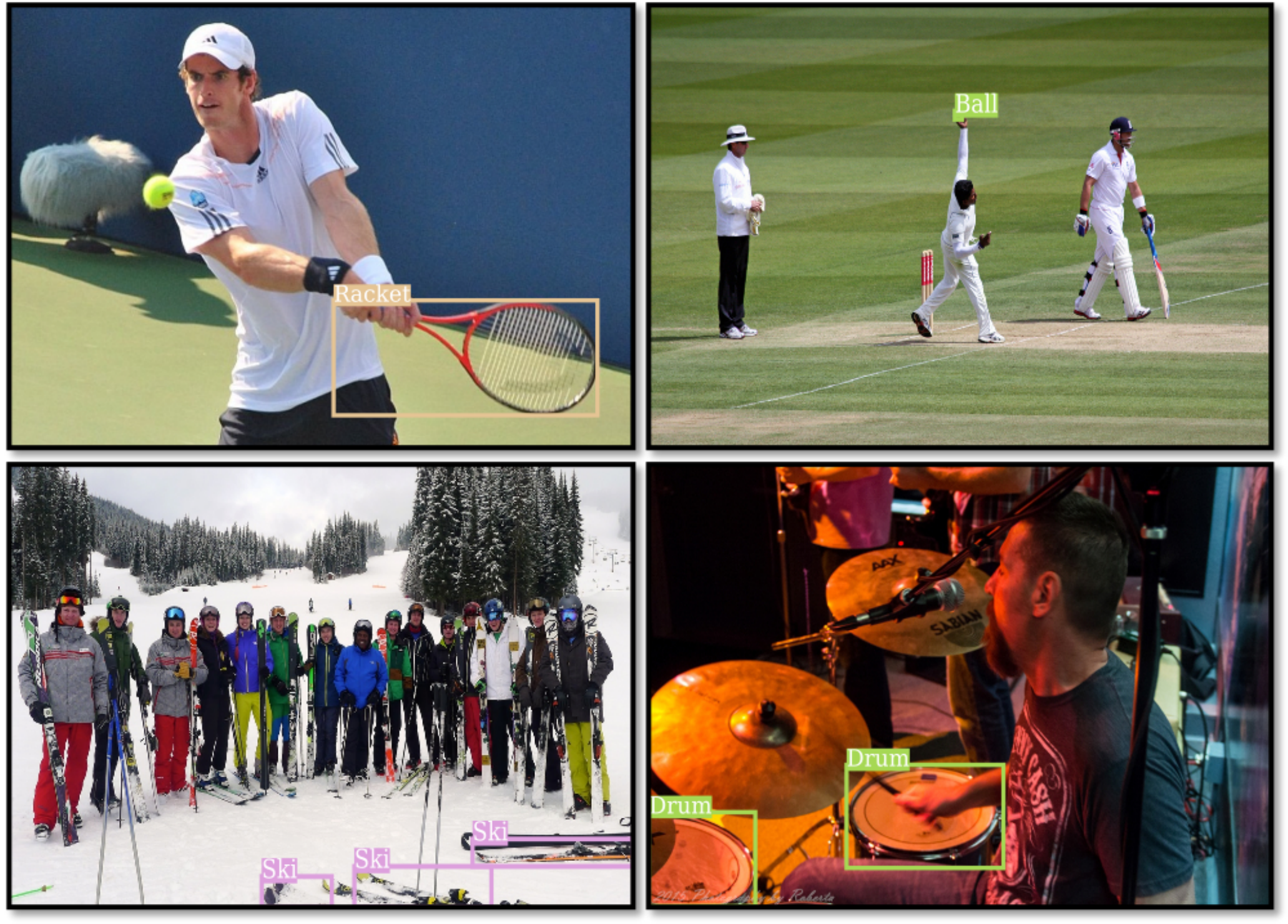}
\end{center}
\caption{
Example images of Open Images.
There are severe missed annotations for the bounding box.
In the example images, all human bounding boxes are missing.
}
\label{fig:weakly_human}
\end{minipage}
\end{figure}

\begin{figure}[t]
\begin{minipage}{\linewidth}
\begin{center}
\includegraphics[width=\linewidth]{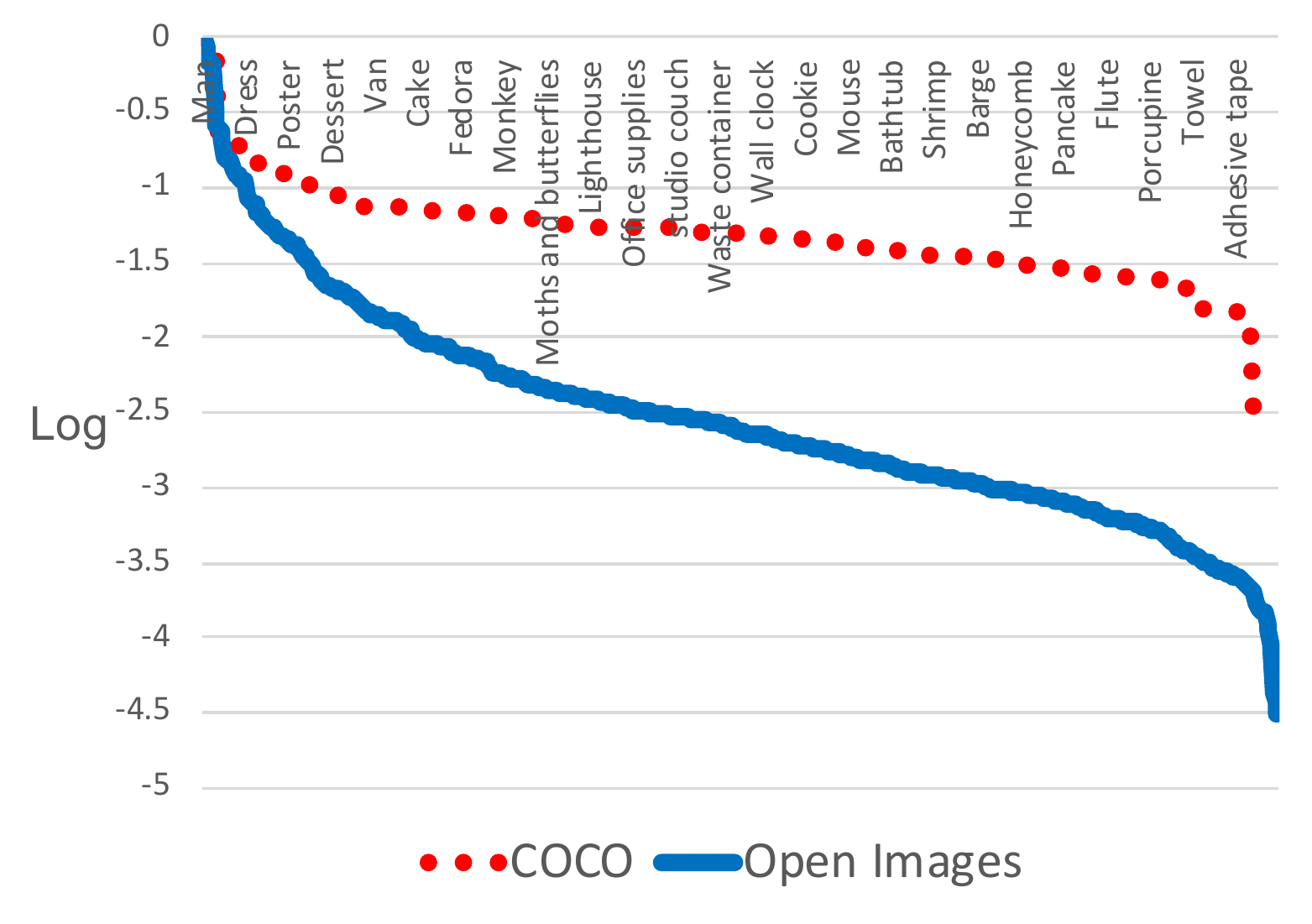}
\end{center}
\caption{
The data statistics of the COCO and the Open Images datasets.
The x-axis and the y-axis are with the categories and the log transformed instance counts, respectively.
It should be noted that the category number of the Open Images and the COCO datasets are different, which is 500 and 80, respectively.
For better visualization, we have aligned the statistics of COCO and Open Images datasets in x-axis.
}
\label{fig:imbalance}
\end{minipage}
\end{figure}

In the era of deep learning, more training data always benefits the generalizability of model~\cite{mahajan2018exploring}.
With the help of large-scale Open Images detection dataset, the frontier of object detection would be pushed forward a great step.
To take full advantage of the large-scale data, we employ EfficientNet~\cite{tan2019efficientnet} as our backbone.
We first grid search an EfficientNet-B1 on Open Images Detection Dataset and then scale it up based on compound scaling method.
However, it is found that the standard compound scaling method~\cite{tan2019efficientnet} is not optimal in the setting of multi-scale training and testing pipeline which is a common data augmentation strategy for object detection.
We argue that it is because that the standard EfficientNet scale-up the resolution to achieve better performance in single-scale training, but it is harmful when training and testing model in multi-scale input.
To remedy this problem, we fix the resolution and re-assign the stage of EfficientNet-B7 to mimic the scheme how ResNeXt~\cite{xie2017aggregated} assigns different amount parameters to different stages architecture.

The hierarchical tag system also should be taken into consideration.
Open Images dataset contains 500 categories, consisting of 5 different levels and 57 parent nodes.
And in the evaluation procedure, a parent category is asked to be output if its child present.
Besides, It is observed that some categories are ambiguous.
For example, almost every \textit{Torch} instance highly overlaps with a \textit{Flashlight} bounding box, although there is not a hierarchical relation between them.
Considering these fact, We propose distributed softmax loss to replace the standard softmax loss.

The large volume of images and categories of Open Images dataset limits its labeling style.
As the enormous numbers of images and categories, it is practical impossible to annotate every instance in every image exhaustively,
there are severe missed annotations for the bounding box, as demonstrated in Figure~\ref{fig:weakly_human}.
Hence, the missed annotations turns the Open Images Detection task into weakly supervised scenario with label noise.
We also observe that Open Images dataset suffers from severe data imbalance, as shown in Figure~\ref{fig:imbalance}.
For example, the category of \textit{Person} has 1.4 million instances, which is 105 times larger than that of the \textit{Pressure cooker} which only has 14 instances.
we propose the class-aware sampling to make a balance between major and rare categories.
Furthermore, to alleviate the overfitting caused by class-aware sampling, we apply auto augmentation~\cite{zoph2019learning} on both image-level and box-level data.
Alternatively, we also train expert model~\cite{akiba2018pfdet, niitani2019sampling} on rare categories then ensemble them to solve the data imbalance problem.

In conclusion, our method has the following advantages:
\begin{itemize}
\item We present an variant of EfficientNet, which overcomes the ineligibility in multi-scale training and testing for object detection.
\item We propose a distributed softmax loss that generalizes the standard softmax loss by exploiting category relationship.
It is more friendly to the hierarchical tag system and more robust to the label noise.
\item We propose class-aware sampling along with auto augmentation to solve the severe data imbalance.
We also present an effective method to train expert model for data imbalance.
\end{itemize}

\section{Methods}
In this section, we will detailedly demonstrate our methods used in this challenge.

\subsection{EfficientNet for Object Detection}

In this work, the Faster R-CNN framework is adopted as our baseline, where the backbone is EfficientNet.
EfficientNet is recent state-of-the-art in many classification tasks.
But there is still limited research about successfully applying EfficientNet to the field of object detection.

EfficientNet is proposed as a paradigm to scale-up Convolutional Neural Networks (CNN) in three dimensions of network, \ie, depth, width, and resolution.
In contrary to previous work, EfficientNet balances these three dimensions by grid search their scale-up coefficient simultaneously.
The grid search is applied in a small search space, then the final network which lays in a larger search space will be sampled by using the compound scaling method.
As there is no pretrained EfficientNet for object detection, we search it directly on a subset of Open Images dataset.
Firstly, we trained a basic model which is so-called EfficientNet-B0.
Then we grid search the next level EfficientNet-B1 whose scale-up coefficient equals to 2.
This depth, width, and resolution of EfficientNet-B1 can be generated by:
\begin{equation}
\label{eq:b2}
\begin{aligned}
d, w, r & = \mathop{\arg\max}_{\alpha, \beta, \gamma}( mAP(model(\alpha, \beta, \gamma))) \\
& \text{s.t.} \ \ \ {\alpha} \cdot {\beta}^{2} \cdot {\gamma}^{2} \approx 2
\end{aligned},
\end{equation}
where $d, w, r$ are the optimal depth, width, and resolution of EfficientNet-B1, respectively.
And the $\alpha, \beta, \gamma$ are temporary depth, width, and resolution in grid search, respectively.
The $model$ is the function that generates and trains detection model according to its input depth, width, and resolution, which is then evaluated by $mAP$ function.
After building EfficientNet-B1, we further obtain EfficientNet-B7 by using the compound scaling method:
\begin{equation}
\label{eq:b7}
\begin{aligned}
\text{EfficientNet-B7} = model({d}^{7}, {w}^{7}, {r}^{7})
\end{aligned}.
\end{equation}

\begin{figure}[t]
\begin{minipage}{\linewidth}
\begin{center}
\includegraphics[width=0.95\linewidth]{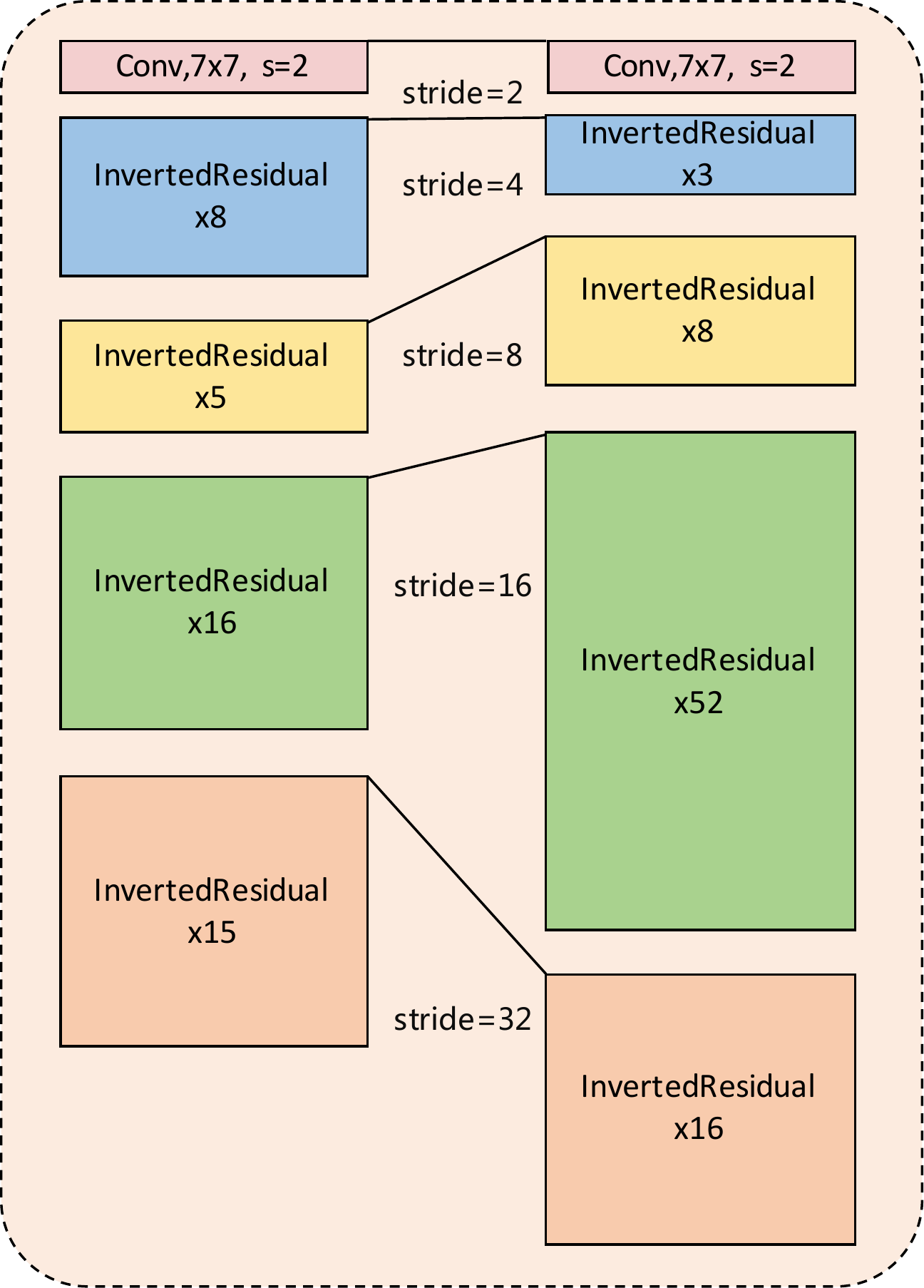}
\end{center}
\caption{
Left: The standard EfficientNet-B7. Right: The proposed variant of EfficientNet-B7.
We put more convolutional layers in stage four, so that this architecture could achieve better performance in a specific input scale.
}
\label{fig:nb7}
\end{minipage}
\end{figure}

Although EfficientNet-B7 achieve better performance in single-scale training and testing.
But under the multi-scale training and testing, it is found that EfficientNet-B7 is slightly inferior to the ResNeXt-152.
We argue that it is because that the standard grid search and compound scaling method is sub-optimal in multi-scale input setting.
EfficientNet carefully assigns depth, width, and resolution to achieve better for objects at all sizes.
However, improving performance for objects at different size is unnecessary even harmful for multi-scale input, because different size objects could be detected better with different input scale.
Thus, we propose to fix the resolution coefficient and only adjust the width and depth.
We further add more blocks into stage four which mimics the ResNeXt-152 to gain the advantage to a specific input scale.
The proposed variant of EfficientNet-B7 is illuminated in Figure~\ref{fig:nb7}.

\subsection{Distributed Softmax Loss}
Open Images dataset contains 500 categories, consisting of 5 different levels and 57 parent nodes.
mAP is evaluated for each of the 500 categories.
For a leaf category in the hierarchy, AP is computed normally.
However, in order to be consistent with the meaning of a non-leaf class, its AP is computed involving all its ground-truth object instances and all instances of its children.
Thus, detector are usually asked to output two even more detected boxes for a single region proposal.

Besides, there are many data noise caused by ambiguous categories in Open Images dataset, such as \textit{Torch} and \textit{Flashlight},
they are so similar in semantics or appearance that the annotators hardly discriminate them and label them as each other.
It is benefit to boost performance if we can output both ambiguous categories when one of them presents.

Unfortunately, the standard softmax cross-entropy loss is arduous to deal with these hierarchical tag system and the data noise.
The standard softmax cross-entropy loss is presented as:
\begin{equation}
\label{eq:standardsoftmax}
\begin{aligned}
\mathcal{L}_{cls} & = \sum_{c=1}^{C} \mathbbm{1}_{y_c=1} log(\frac{e^{x_c}}{\sum_{i=1}^{C}e^{x_i}}) \\
\end{aligned},
\end{equation}
where $C$ is the number of category, $y_c$ is the c-th elements in label vector $\textbf{y}$, and $x_i$ is the i-th elements in logit vector $\textbf{x}$.
It should be noted that the label $\textbf{y}$ is an one-hot vector which only allows one positive output, thus, is not suitable for multi-label classification.
An intuitive solution is to use the binary cross-entropy loss~\cite{durand2019learning}, but in our experiments it can not achieve comparable results.
We also try focal loss~\cite{lin2018focal} upon the binary cross-entropy loss, but still worse than standard softmax loss.
To overcome this problem we propose the distributed softmax loss.
It is formulated as:
\begin{equation}
\label{eq:distributedsoftmax}
\begin{aligned}
\mathcal{L}_{cls} & = \sum_{c=1}^{C} y_c log(\frac{e^{x_c}}{\sum_{i=1}^{C}e^{x_i}}) \\
\end{aligned},
\end{equation}
where $y_c$ belongs to the label vector $\textbf{y}$ with $k$ non-zero elements each set to $1/k$ corresponding to the $k \geq 1$.
The proposed distributed softmax loss not only allows multi-label training as binary cross-entropy loss, but also persists the suppression between categories as standard softmax loss.
In our experiments, it improves the baseline by 1 points of mAP.

\begin{table*}
\small
\centering
\begin{tabular}{lllllll}
\toprule
     & Operation 1 & P  & M    & Operation 2 & P  & M    \\
\midrule
Sub-policy 1 & TranslateX\_BBox & 0.6 & 4 & Equalize & 0.8 & 10            \\
Sub-policy 2 & TranslateY\_Only\_BBoxes & 0.2 & 2 & Cutout & 0.8 & 8       \\
Sub-policy 3 & Sharpness & 0.0 & 8   & ShearX\_BBox & 0.4 & 0              \\
Sub-policy 4 & ShearY\_BBox & 1.0 & 2 & TranslateY\_Only\_BBoxes & 0.6 & 6 \\
Sub-policy 5 & Rotate\_BBox & 0.6 & 10 & Color & 1.0 & 6                   \\
\bottomrule
\end{tabular}
\caption{
The auto augmentation strategies used in our experiments.
}
\label{tab:autoaug}
\end{table*}
\subsection{Class-aware Sampling and Expert Model}
In Figure~\ref{fig:imbalance}, we illuminate the data statistics of the COCO and Open Images datasets.
It can be seen that Open Images dataset exhibits much severe data imbalance compared to the its counterpart.
Following Gao \etal~\cite{gao2018solution}, we apply class-aware sampling to alleviate this problem.
In details, the class-aware sampling firstly samples one category out of 500 uniformly.
And then an image containing objects with the same category of the first step is sampled.
This strategy could assure all the category has the same chance to be trained.

However, overfitting is introduced unavoidably by this class-aware sampling.
We employ auto augmentation in both image level and box level to remedy this problem.
The auto augmentation strategies is listed in Table~\ref{tab:autoaug}.

Expert model is another method that can overcome the data imbalance problem.
In this challenge, we fine-tune the baseline model on a small subset of the full category space to enhance the recall of rare categories.
The baseline model is trained on the entire dataset.
Each expert model is concentrated to a expert subset, for instance, 50 categories out of 500.
Expert model is useful to increase the recall of rare categories, but tends to bring more false positives.
We further develop three strategies to solve the unwanted false positives.
First, we build a confusion matrix to find those categories that is easy to be classified into expert subset.
In next turn training, those confusion categories is added into the training set, which helps the expert model to discriminate the false positives.
Second, we train many expert models with different sizes of expert subset, and there is an overlap between each expert subset.
Only the detected boxes of overlapped categories is considered as final results.
Third, we train a classifier to re-weight the confidence of detected boxes by multiple the detector score and classifier score.
It should be noted that there is not an absolute positive correlation between the mean accuracy of classifier and the final mAP of re-weighted detection results.
We also find that the classifier is beneficial to the normal model that trained on the entire dataset.

\subsection{Ensemble}
For the final results, we have used normal models and expert models for ensemble.
Since the distribution of the performance over categories is different among models, we re-weight each categories model by model according to its mAP on validation set.
In details, we set a weight $w_{c}^{m}$ for model $m$ and its categories $c$.
The $w_{c}^{m}$ is defined as:
\begin{equation}
\label{eq:nms1}
\begin{aligned}
w_{c}^{m} & = \frac{s_{c}^{m} - {\mu}_{c}}{t_c - {\mu}_c} + \alpha \frac{t_{c} - s_{c}^{m}}{t_c - {\mu}_c} \\
\end{aligned},
\end{equation}
where $s_{c}^{m}$ is the validation mAP of the model $m$ for category $c$, ${\mu}_{c}$ and $t_{c}$ is the mean and max mAP of the for category $c$ among all models, respectively.
$\alpha$ is the lower bound of $w_{c}^{m}$ so that if $s_{c}^{m}$ is lower than ${\mu}_{c}$, then $w_{c}^{m}$ is set to $\alpha$.

We also search optimal NMS threshold for different categories by:
\begin{equation}
\label{eq:nms2}
\begin{aligned}
H_c & = \mathop{\arg\max}_{h_c \in (0,1)} (mAP(h_c) + \frac{1}{(h_c - d)^2}) \\
\end{aligned},
\end{equation}
where $h_c$ is the NMS threshold for category $c$, and the $d$ is the default NMS threshold.
$mAP$ is a function mapping NMS threshold to ensemble mAP.
We expect to maximize the validation mAP by searching the NMS threshold, meanwhile, minimize the variance of the NMS threshold.

\section{Experiments}

\subsection{Data}
We train our model on Open Images dataset.
We used the recommended train and validation splits of Open Images Challenge 2018 as its validation set is labelled denser than that of Challenge 2019.
In addition to the Open Images dataset, Objects365~\cite{obj365} is used to train the final models for categories that intersects with Open Images dataset.

\subsection{Results}
\begin{table}
\small
\centering
\begin{tabular}{ll}
\toprule
Method & Public Leader board    \\
\midrule
Baseline (FPN with ResNeXt-152) & 53.88 \\
+EfficientNet-B7                & 55.59 \\
+Distributed Softmax Loss       & 56.43 \\
+Class-aware Sampling           & 61.09 \\
+Auto Augmentation              & 61.84 \\
+Classifier                     & 62.29 \\
+Ensemble                       & 67.17 \\
\bottomrule
\end{tabular}
\caption{
Performance improvement by adding different strategies
step by step.
}
\label{tab:results}
\end{table}
During the training process, we follow the typical hyper-parameter settings.
We use SGD with momentum 0.9 and weight decay 0.0001.
The initial learning rate is set to 0.00125 per batch, and warm-up~\cite{goyal2017accurate} phase is adopted.
All models are trained from scratch~\cite{he2018rethinking}.

Table~\ref{tab:results} demonstrates the detailed ablation with our method.
As can be seen, the variant of EfficientNet-B7 surpasses the ResNeXt-152 by 1.71\%.
After solving the hierarchical tag system and data noise by distributed softmax loss, the performance is boosted by 0.84 points.
The class-aware sampling balances the dataset and leads to a heavy gain, \ie, 4.67 points.
By further using auto augmentation and classifier, the model is improved by  0.75\% and 0.45\%, respectively, achieving the best single model with a mAP of 62.29\%.
With a final ensemble of 12 different models, we achieve the 67.17\% mAP in the public leaderboard and 64.21\% mAP in the private leaderboard, ranking the 3rd place.

\section{Conclusions}
In this work, we present an variant of EfficientNet for detection which is suitable for multi-scale training and testing.
On the top of efficient backbone, we introduces a novel distributed softmax loss for the hierarchical label and data noise,
and use class-aware sampling, auto augmentation, expert model and classifier to solve the data imbalance.
The experiments demonstrate the effectiveness of our method.

{\small
\bibliographystyle{ieee_fullname}
\bibliography{egbib}
}

\end{document}